\documentclass{article}

\usepackage[accepted]{icml2025}

\usepackage{microtype}
\usepackage{graphicx}
\usepackage{subcaption}
\usepackage{booktabs} 

\usepackage{xcolor}
\usepackage{xspace}
\usepackage{amsmath, amssymb}
\usepackage{xcolor}
\usepackage[absolute,overlay]{textpos}
\usepackage[capitalize]{cleveref}

\usepackage{algorithm}
\usepackage{algorithmic}

\newcommand{\algoNameFull}{S$^2$Edit\xspace}
\newcommand{\heading}[1]{\noindent\textbf{#1}}

\icmltitlerunning{\algoNameFull}

\crefname{section}{Sec.}{Secs.}
\Crefname{section}{Section}{Sections}
\Crefname{table}{Table}{Tables}
\crefname{table}{Tab.}{Tabs.}
\Crefname{figure}{Figure}{Figures}
\crefname{figure}{Fig.}{Figs.}

\begin{document}
\begin{textblock*}{\paperwidth}(0cm, 1cm) 
    \centering
    \textcolor{gray}{\small This paper has been accepted for publication by the ICML 2025 workshop on New In ML}
\end{textblock*}
\twocolumn[
\icmltitle{\algoNameFull: Text-Guided Image Editing with Precise Semantic and Spatial Control}

\begin{icmlauthorlist}
\icmlauthor{Xudong Liu}{modiface,ut}
\icmlauthor{Zikun Chen}{modiface}
\icmlauthor{Ruowei Jiang}{modiface}
\icmlauthor{Ziyi Wu}{ut}
\icmlauthor{Kejia Yin}{modiface,ut}
\icmlauthor{Han Zhao}{uiuc}
\icmlauthor{Parham Aarabi}{modiface,ut}
\icmlauthor{Igor Gilitschenski}{ut}
\end{icmlauthorlist}

\icmlaffiliation{ut}{Department of Computer Science, University of Toronto}
\icmlaffiliation{modiface}{ModiFace Inc.}
\icmlaffiliation{uiuc}{Department of Computer Science, University of Illinois Urbana-Champaign}

\icmlcorrespondingauthor{Xudong Liu}{liuxd1215@gmail.com}
\icmlcorrespondingauthor{Igor Gilitschenski}{gilitschenski@cs.toronto.edu}

\icmlkeywords{Image Editing, Diffusion Models}

\vskip 0.3in]

\printAffiliationsAndNotice{} 

\begin{abstract}
Recent advances in diffusion models have enabled high-quality generation and manipulation of images guided by texts, as well as concept learning from images. However, naive applications of existing methods to editing tasks that require fine-grained control, \textit{e.g.}, face editing, often lead to suboptimal solutions with identity information and high-frequency details lost during the editing process, or irrelevant image regions altered due to entangled concepts. In this work, we propose \algoNameFull, a novel method based on a pre-trained text-to-image diffusion model that enables personalized editing with precise semantic and spatial control. We first fine-tune our model to embed the identity information into a learnable text token. During fine-tuning, we disentangle the learned identity token from attributes to be edited by enforcing an orthogonality constraint in the textual feature space. To ensure that the identity token only affects regions of interest, we apply object masks to guide the cross-attention maps. At inference time, our method performs localized editing while faithfully preserving the original identity with semantically disentangled and spatially focused identity token learned. Extensive experiments demonstrate the superiority of \algoNameFull over state-of-the-art methods both quantitatively and qualitatively. Additionally, we showcase several compositional image editing applications of \algoNameFull such as makeup transfer.
\end{abstract}

\section{Introduction}
\label{sec:intro}

Recent years have witnessed remarkable progress in the field of generative models, with techniques such as Generative Adversarial Networks (GANs)~\cite{goodfellow2014generative,karras2019style,karras2020analyzing,karras2021alias} and Diffusion Models~\cite{sohl2015deep,saharia2022photorealistic,unCLIP,rombach2022high} capable of synthesizing realistic images.
These advancements have fueled significant interest in the realm of image editing, a highly practical and versatile domain.
Among its various settings, text-guided editing stands out for its user-friendly interface, as it accepts natural language as the input.

Early methods in this field often train GANs from scratch on paired image-text data~\cite{GANEdit1,GANEdit2,GANEdit3}, requiring expensive annotations.
Later approaches explore the latent space of pre-trained models like CLIP~\cite{radford2021learning}, which align text and image features.
However, GANs are still trained on relatively small datasets, resulting in a latent space with limited capacity.
They sometimes struggle to represent real-world objects at the tails of distributions faithfully, and fail to generalize to unseen text prompts.
Recent advances in large-scale pre-trained text-to-image diffusion models~\cite{saharia2022photorealistic, rombach2022high} offer alternative solutions to this task.
Thanks to the large scale pre-trained data, these methods exhibit a significantly stronger capacity in representing diverse objects from different domains, and are generally more robust to text prompt variations.
Nevertheless, demonstrations of these editing methods' performance typically involve marked changes in art styles, object shapes, motions, and precise editing of real-world images such as faces remains under-explored.
When guided by detailed prompts, these methods frequently introduce severe artifacts or unrealistic styles. 
Moreover, they sometimes fail to edit the image utterly, with object identity information changed or the target attribute indicated in the prompt missing in the output.
Essentially, current methods struggle to balance between identity information preservation and target prompt alignment~\cite{SINE,kawar2023imagic, brooks2023instructpix2pix}, and significant challenges still remain in editing real-world images both accurately and faithfully.

In response to these challenges, we propose \algoNameFull, a two-stage text-guided image editing method.
The first stage involves fine-tuning a pre-trained diffusion model to reconstruct the input image conditioned on a text prompt, which helps capture the object identity.
To apply target edits while keeping the original identity information intact, we insert a learnable identity token into the prompt to capture nuanced identity features.
However, plain fine-tuning lacks the control of the semantic meaning or spatial emphasis of the identity token, which is critical for precise image editing.
We observe that the learned identity token often contains information on attributes targeted for modification, which impedes the later editing process.
To disentangle the identity token from target attributes, we apply semantic control by enforcing an orthogonality constraint between the identity token and the prompt in the textual feature space during fine-tuning.
Moreover, the identity token may learn to represent identity-irrelevant image regions, undermining the preservation of identity information.
Therefore, we apply spatial control during fine-tuning by manipulating the cross-attention maps of the identity token with an object mask.
In the second stage of \algoNameFull, we augment the cross-attention injection strategy from Prompt-to-Prompt~\cite{hertz2022prompt} with spatial control, and employ it for editing with our identity token.
With its high-quality editing results, \algoNameFull outperforms state-of-the-art methods both quantitatively and qualitatively in evaluations, and garners at least 45\% more preference from the participants in a user study.

In summary, we make the following main contributions:
\begin{enumerate}
    \item We present \algoNameFull, a novel text-guided image editing method that excels at obtaining fine-grained controls over local details.
    \item To balance identity preservation and prompt alignment in image editing, we enhance the personalized fine-tuning approach to control semantic meaning and spatial focus of the learnable identity token.
    \item \algoNameFull achieves superior editing results over previous methods on multiple datasets. Moreover, we apply it to compositional image editing tasks like makeup transfer.
\end{enumerate}
\section{Related Work}\label{sec:relatedworks}
\heading{Diffusion-based Image Editing.}
First introduced by Sohl-Dickstein \textit{et al.}~\cite{sohl2015deep}, diffusion models have recently achieved state-of-the-art image synthesis quality~\cite{ho2020denoising,dhariwal2021diffusion,nichol2021improved,saharia2022photorealistic,rombach2022high}.
Based on these models, image editing works perform tasks like inpainting~\cite{song2020score, paint-by-example}, stroke-based editing~\cite{meng2021sdedit,BlendedDMEdit,BlendedLDMEdit}, and text-guided editing~\cite{hertz2022prompt,PnP-Diffusion,parmar2023zero}.
In particular, text-guided editing garners considerable interest due to its friendly interface and substantial control capabilities.
Studies in this domain enable rich text guidance by manipulating attention maps~\cite{hertz2022prompt,PnP-Diffusion,parmar2023zero}, text prompts~\cite{nichol2021glide,mokady2023null,brooks2023instructpix2pix,kawar2023imagic,pan2023effective,PromptTuningInversion}, and feature space~\cite{kim2022diffusionclip,bansal2023universal}.
Prompt-to-Prompt~\cite{hertz2022prompt} investigates the relation between prompt tokens and spatial layout of images, and edits images by manipulating the cross-attention layers.
InstructPix2Pix~\cite{brooks2023instructpix2pix} fine-tunes a pre-trained model on synthesized paired images and editing instructions.
Despite impressive results, Prompt-to-Prompt sometimes fails to balance identity preservation and desired editing effects, and its performance often depends on the quality of the given prompt.
InstructPix2Pix partially addresses the second issue with more straightforward instruction guidance, yet the first issue remains as it does not provide identity-related control.

\heading{Personalized Image Synthesis and Manipulation.}
To preserve the fidelity of the original image, recent works in the image diffusion domain propose to insert learnable tokens and fine-tune pre-trained text-to-image models such that identity information can be learned and re-combined with new contexts~\cite{gal2022image,kumari2023multi,ruiz2023dreambooth,shi2023instantbooth}.
DreamBooth~\cite{ruiz2023dreambooth} learns to encode object identity in a unique token by fine-tuning diffusion models on several images containing the same object.
SINE~\cite{SINE} investigates the overfitting problem when fine-tuning involves only one image, and proposes to distill the knowledge of the fine-tuned model into the original model.
Instead of fine-tuning the entire model, Custom Diffusion~\cite{kumari2023multi} only optimizes the Key and Value projection layers in cross-attention, and is able to compose multiple concepts via joint training on images with various objects.
Although personalized fine-tuning helps preserve object identity, we observe a lack of semantic and spatial control over the learned identity token in existing methods, leading to failure modes such as entangled concepts when applied to detailed editing tasks.

\heading{GAN-based Image Editing.}
One line of work on GAN attempts to discover meaningful feature space directions~\cite{harkonen2020ganspace,shen2020interfacegan,chen2022exploring,wu2021stylespace} that correspond to different semantic changes of images.
Some other methods invert images to the latent space of pre-trained GANs~\cite{richardson2021encoding,tov2021designing,wang2022high,lyu2023deltaedit,DM-GAN-Edit} and leverage the aforementioned semantic directions for editing.
This direction is further spurred by the development of large vision-language models such as CLIP~\cite{radford2021learning}.
DeltaEdit~\cite{lyu2023deltaedit} proposes a CLIP delta space where visual and textual features from CLIP are better aligned, and trains a mapping network that connects the delta space to StyleGAN's~\cite{karras2020analyzing} latent space to enable text-guided editing.
However, compared to state-of-the-art diffusion models, GANs, which are trained on smaller datasets, have a more limited expressibility in their latent space.
\section{Method}\label{sec:method}

\algoNameFull builds upon state-of-the-art text-to-image diffusion models and comprises two stages: The identity fine-tuning stage and the inference stage.
During fine-tuning, we learn a special identity token to preserve the subtle information in the original image, and propose novel semantic and spatial controls to enable accurate and faithful editing.
In the inference stage, we insert the learned special token during fine-tuning into any target prompt.
In addition, we extend \algoNameFull to the task of compositional image editing. For preliminaries of text-guided image editing, please refer to the Appendix~\textcolor{red}{A.1}.
\begin{figure}[t]
    \centering
    \includegraphics[width=1.0\linewidth]{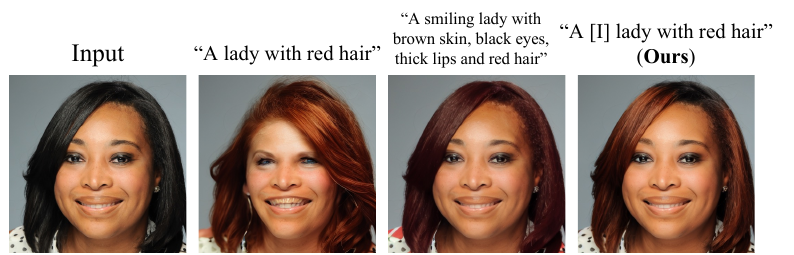}
    \vspace{-1mm}
    \caption{
    Impact of prompts on editing results.
    The results are guided by the prompts listed above.
    }
    \label{fig:method-shortlong}
    \vspace{-4mm}
\end{figure}

\begin{figure*}[t]
    \centering
    \includegraphics[width=1.0\linewidth]{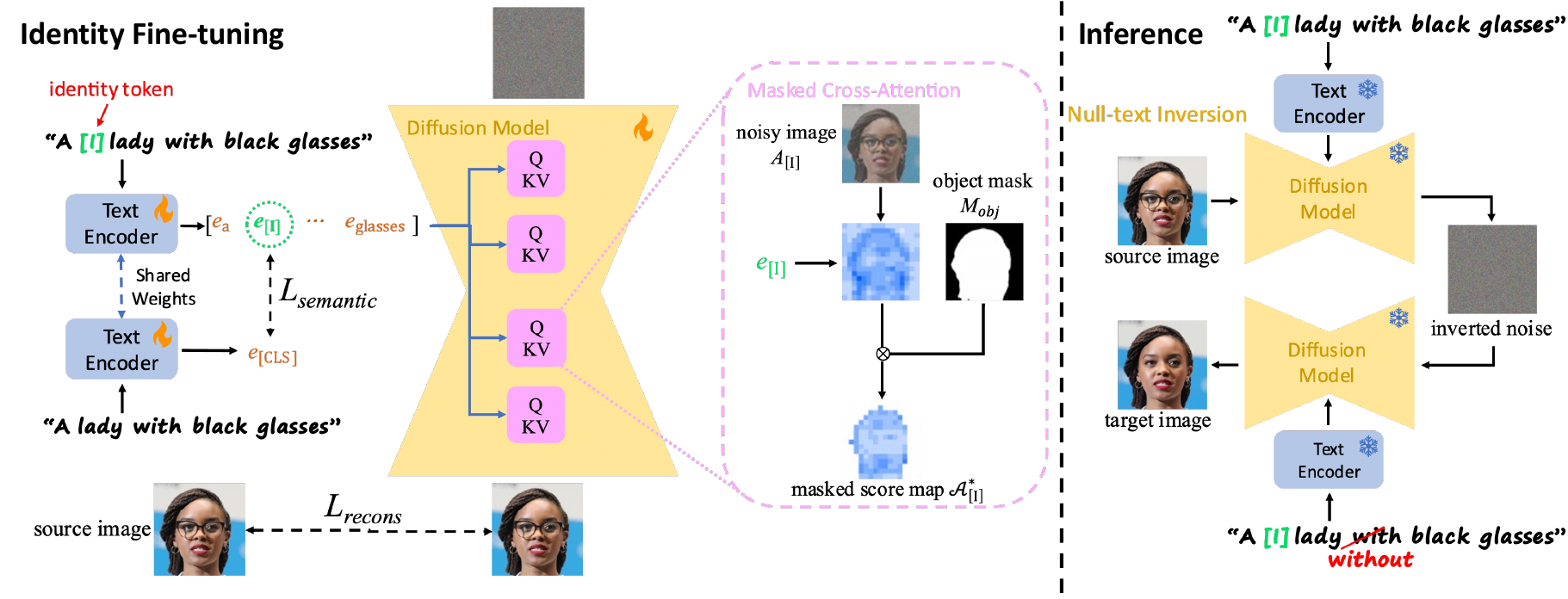}
    \caption{
        \textbf{\algoNameFull overview.}
        \textit{Left}: Given a source image and a text prompt, we insert a learnable token [I] into the text prompt and fine-tune a pre-trained text-to-image diffusion model to learn the identity information.
        To obtain a disentangled identity token, we apply an orthogonality constraint in the text embedding space via a semantic loss $L_{semantic}$ and force [I] to only represent the object of interest with masked cross-attention.
    \textit{Right}: With [I] learned, we freeze the fine-tuned model and perform Null-text Inversion~\cite{mokady2023null} to get an initial noise map, then denoise it conditioned on the target prompt
        to generate the editing result.
    }
    \label{fig:pipeline}
    \vspace{-3mm}
\end{figure*}

\subsection{Identity Fine-tuning}
Although plain image editing following procedures as described above can edit images according to the prompt, in our preliminary experiments, it often fails to preserve subtle details like identity-related details for face editing. 
\cref{fig:method-shortlong} shows that this may originate from unclear guidance of the prompts.

Instead of optimizing the prompt in the text space, we adapt the special token introduced in DreamBooth~\cite{ruiz2023dreambooth} to our method by constructing enhanced prompts $\widetilde{\mathcal{P}}$ and $\widetilde{\mathcal{P}^\ast}$ with a unique identity token [I] (\textit{e.g.} ``a [I] man with beard''), which encodes rich identity information and provides clear guidance to the generation process. Different from DreamBooth, which aims to extract appearance of a subject from multiple images for recontextualization tasks, we learn the identity information from one image to ensure that subtle details can be flawlessly preserved.
We fine-tune the diffusion model and its text encoder with $\mathcal{I}$ and the enhanced prompt $\widetilde{\mathcal{P}}$ to learn the identity information, which is encoded into both the embedding of [I] and the model weights.

\subsection{Precise Semantic and Spatial Control}\label{sec:s2-control}
Plain fine-tuning facilitates the generation of high-fidelity images with the original identity preserved, yet it lacks control over information encoded in the identity token [I], resulting in prompt alignment failures like concept entanglement or omission of target attributes.
In this section, we propose semantic and spatial control to balance between identity preservation and prompt alignment.

\heading{Semantic control of the identity token.}
One issue of plain identity fine-tuning is that the identity token [I] may learn to represent attributes targeted for modification from the source image $\mathcal{I}$.
This is problematic as we can hardly manipulate information encoded in [I] after fine-tuning, since the learned information is implicitly stored in both text embeddings and the model weights.
To address this issue, we build upon the insight that users always specify the attributes they want to edit in the source prompt $\mathcal{P}$.
Therefore, we can disentangle the textual semantic spaces of [I] and $\mathcal{P}$ by adding a semantic loss during fine-tuning:
\begin{equation}
    \mathcal{L}_{semantic} = \|\mathrm{Proj}_{e_{\mathcal{P}}}(e_{[\text{I}]})\| = \|\cos(e_{\mathcal{P}}, e_{[\text{I}]})\|,
\end{equation}
where $e_{[\text{I}]}$ is the text embedding of [I], $e_{\mathcal{P}}$ is the text embedding of the \texttt{[CLS]} token from $\mathcal{P}$, which represents the overall semantic meaning of the entire prompt, and $\cos(e_{\mathcal{P}}, e_{[\text{I}]})$ is the cosine similarity score. 
$\mathcal{L}_{semantic}$ forces $e_{[\text{I}]}$ to be orthogonal to $e_{\mathcal{P}}$, thus reducing their overlaps in the semantic space.
The final objective of identity fine-tuning now becomes:
\begin{equation}
    \mathcal{L} = \mathcal{L}_{recons} + \lambda \cdot \mathcal{L}_{semantic}.
\end{equation}
With a proper loss weight $\lambda$, the learned identity token shares little information with the source prompt $\mathcal{P}$, allowing disentangled and flexible edits guided by the target prompt.

\begin{figure*}[t]
    \centering
    \includegraphics[width=0.8\linewidth]{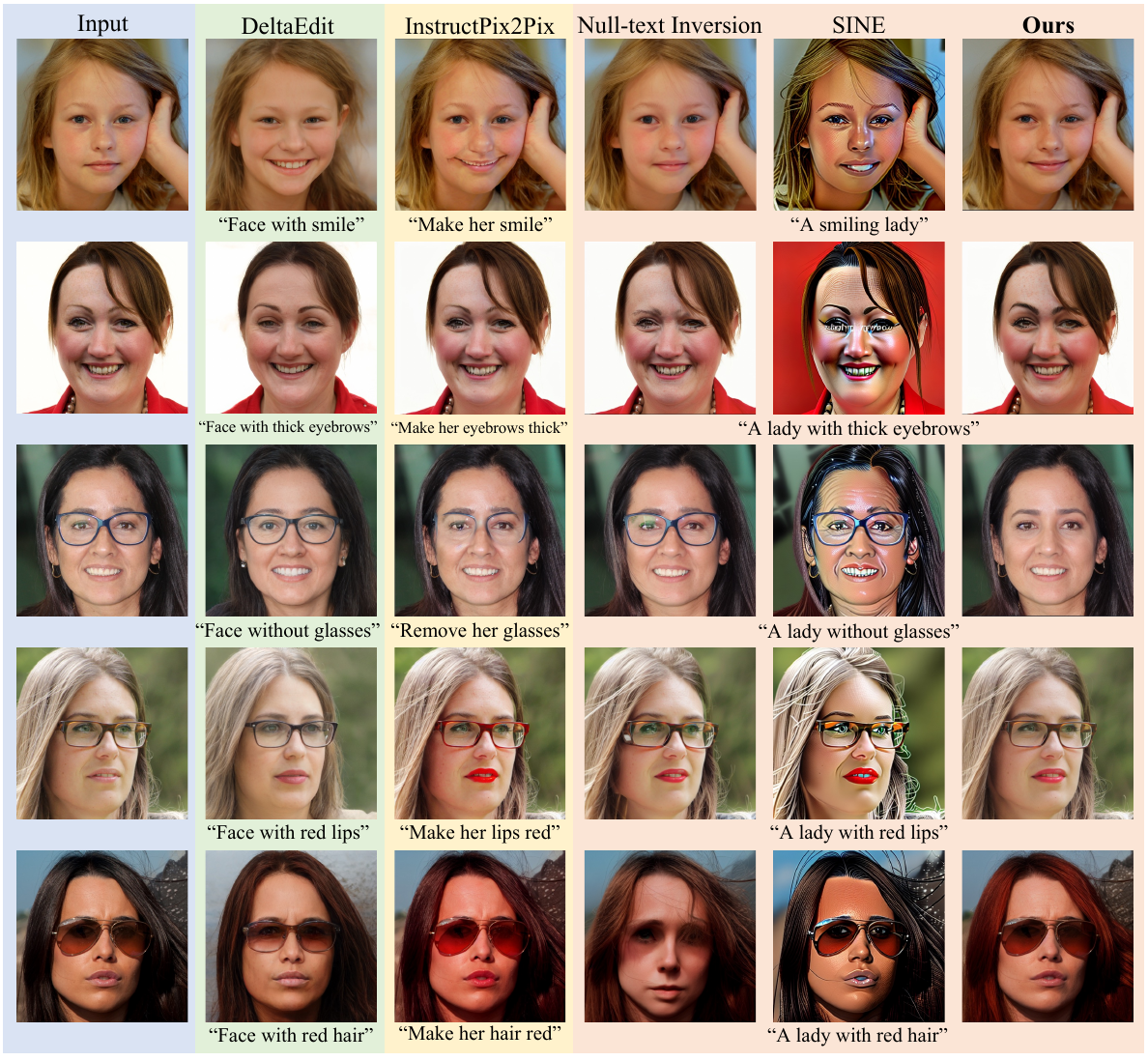}
    \caption{
        Qualitative comparison of text-guided image editing in the face domain.
        The target prompts are listed under each row.
        \algoNameFull outperforms state-of-the-art methods significantly with accurate and faithful edits that align well with the editing prompts while preserving identity information.
        Prompt details are provided in Appendix~\textcolor{red}{A.3}
    }
    \label{fig:compare-baseline}
    \vspace{-4mm}
\end{figure*}

\heading{Spatial control of the identity token.}
Another issue with identity fine-tuning is that the identity token [I] may prioritize less significant image regions, undermining the preservation of the crucial identity information for faithful editing.
To constrain the effective region of [I], we manipulate the cross-attention maps of [I] with an object mask in the denoising U-Net during both fine-tuning and inference stages.
Specifically, we force [I] to only attend to the object the user wants to edit via: 
\begin{equation}
    \mathcal{A}^\ast_{\text{[I]}} = \mathcal{A}_{\text{[I]}} \odot M_{obj},
\end{equation}
where $\mathcal{A}_{\text{[I]}}$ is the original query-key cross-attention map of [I], $\mathcal{A}^\ast_{\text{[I]}}$ is its masked version, $M_{obj}$ is the binary mask of the object of interest, and $\odot$ is element-wise product.
To obtain $M_{obj}$, we take the cross-attention map of the word that describes the object of interest (\textit{e.g.}, ``lady", ``cat") and binarize it, thus requiring no mask annotation.
Compared to other text-guided image editing techniques, such as DeltaEdit, Imagic, SINE, and Prompt Tuning Inversion \cite{lyu2023deltaedit, kawar2023imagic, SINE, PromptTuningInversion}, our approach imposes no additional demands on users, apart from specifying the object within the source prompt.

Overall, with semantic and spatial control, the identity token [I] is disentangled from attributes to be edited and irrelevant image regions.
Thus, it can preserve fidelity to the source image without hindering the editing process.

\subsection{Compositional Image Editing}\label{sec:attr-mixing}

We extend our method to compose attributes across multiple images, \textit{i.e.}, copying attributes from a reference image to a source image.
The inputs of this task are a source image $\mathcal{I}_{src}$ and a reference image $\mathcal{I}_{ref}$, along with their corresponding prompts $\mathcal{P}_{src}$ and $\mathcal{P}_{ref}$, where $\mathcal{P}_{ref}$ contains the attribute we want to copy.
To preserve the identity information of $\mathcal{I}_{src}$, we insert an identity token [I] into $\mathcal{P}_{src}$ to create $\mathcal{P}^\ast_{src}$, \textit{e.g.}, ``a [I] lady''.
To extract the desired attribute from the reference image, we insert an attribute token [A] to $\mathcal{P}_{ref}$ and form $\mathcal{P}^\ast_{ref}$, \textit{e.g.}, ``a model with [A] makeup''.
Next, we fine-tune a diffusion model and its text encoder using two image-prompt pairs, ($\mathcal{I}_{src}$, $\mathcal{P}^\ast_{src}$) and ($\mathcal{I}_{ref}$, $\mathcal{P}^\ast_{ref}$), with semantic and spatial control.
After fine-tuning, the identity and attribute information are encoded into tokens [I] and [A], respectively.
We then combine [I] and [A] into one prompt $\mathcal{P}_{mix}$, \textit{e.g.}, ``A [I] lady with [A] makeup'', and use $\mathcal{P}_{mix}$ to generate the composed image $\mathcal{I}_{mix}$ with the inversed noise map of $\mathcal{I}_{src}$.
\algoNameFull achieves flexible and precise attribute transfers.

\begin{figure*}[t]
  \vspace{2mm}
  \centering
  \includegraphics[width=0.8\linewidth]{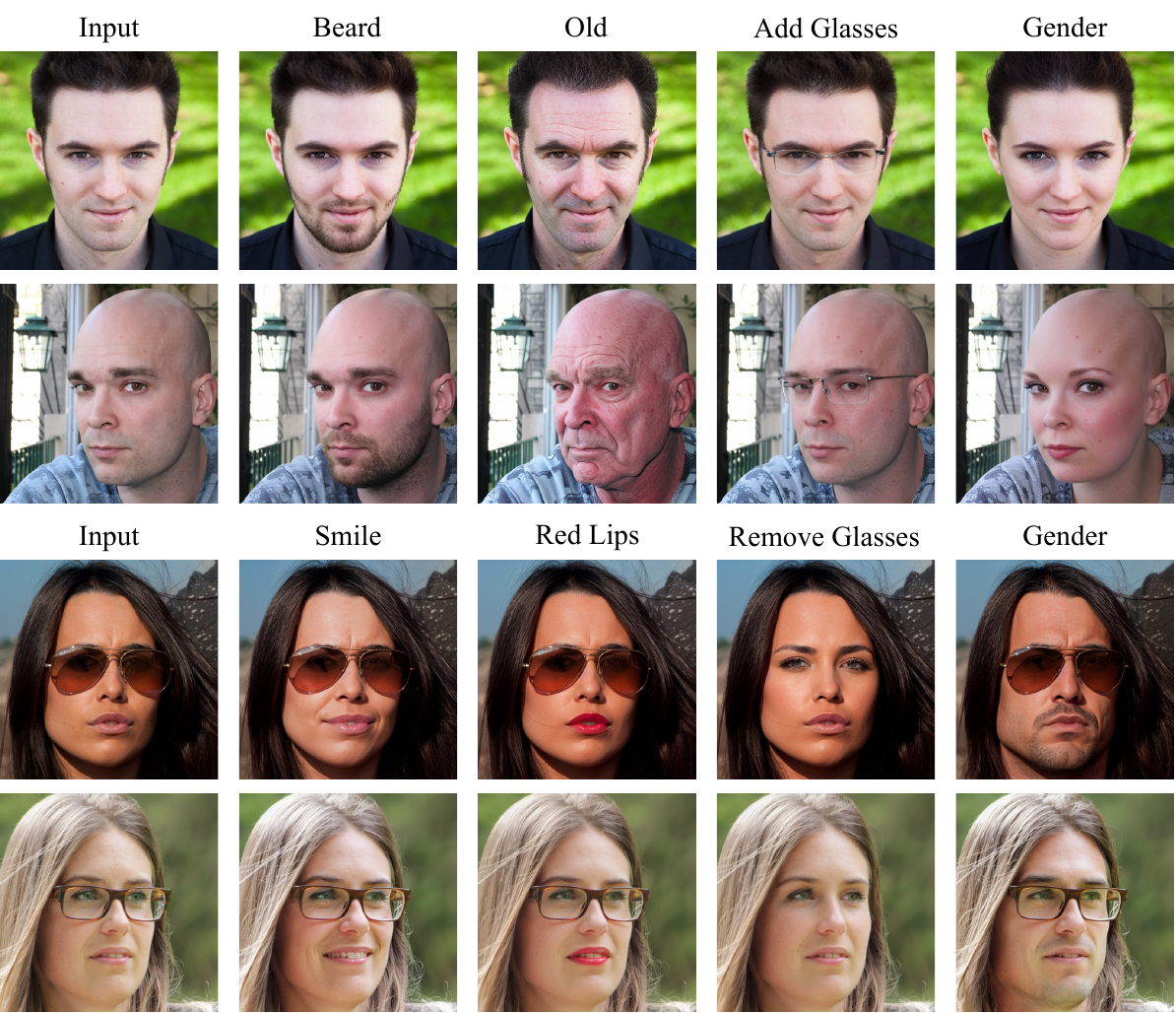}
  \caption{
     Fine-grained editing results of \algoNameFull on the same image for various attributes.
    Full prompts used are provided in Appendix~\textcolor{red}{A.3}.
  }
  \label{fig:same-identity}
  \vspace{-7mm}
\end{figure*}

\section{Experiments}\label{sec:experiment}

In this section, we evaluate our method on text-guided real image editing.
We first introduce our setup in \cref{sec:exp-setup}.
In \cref{sec:qual-results} and \cref{sec:quan-results} we present qualitative and quantitative comparisons with four state-of-the-art methods on several datasets.
Finally, we conduct ablation studies on the components and hyper-parameter of \algoNameFull in \cref{sec:ablations}.

\subsection{Experimental Setup}\label{sec:exp-setup}
\heading{Identity Fine-tuning and Inference.}
We fine-tune a diffusion model and its text encoder with $\mathcal{I}$ and $\widetilde{\mathcal{P}}$ with only one image to learn the identity information, as parameter-efficient fine-tuning results in suboptimal editing performance (see Appendix~\textcolor{red}{B}). 
For detailed cross-attention injection schemes during the inference phase, please refer to Appendix~\textcolor{red}{A.2}.


\heading{Implementation Details.}
While \algoNameFull can be applied to any text-to-image diffusion model, we implement it based on Stable Diffusion v1-4~\cite{rombach2022high} due to its public availability and importance in the literature.
During identity fine-tuning, we use the AdamW~\cite{adamw} optimizer with a base learning rate of $2\times10^{-6}$.
The semantic loss weight $\lambda$ is set to $0.1$.
We fine-tune the model for 200 steps to avoid overfitting. 
For the DDIM scheduler, the number of diffusion steps $T$ is set to 50, and the other hyperparameters are the same as Stable Diffusion~\cite{rombach2022high}. 
All experiments are conducted on one NVIDIA A100 GPU, where fine-tuning takes around 95 seconds, Null-text inversion process takes 113 seconds, and the inference time is 9 seconds.
For the same source prompt and input image, our method requires fine-tuning and inversion to be run once, then only needs 9 seconds to generate edited results with different target prompts.

\heading{Datasets.}
To verify the editing performance and generalization ability of our proposed method, we conduct extensive experiments on images of diverse objects. 
Given the higher sensitivity of humans to detailed alterations and unnatural features in human faces, face editing serves as an ideal experiment for editing performance evaluation. 
Hence, we categorize source images into face and non-face domains, showcasing more results in face editing following \cite{lyu2023deltaedit}.
For the human face domain, we use images from FFHQ~\cite{karras2019style} and CelebA~\cite{celeba}.
For non-face domains, we provide results on AFHQ~\cite{afhq}, LSUN~\cite{lsun} cat and church images.

\heading{Comparisons.}
We compare \algoNameFull with several state-of-the-art diffusion-based text-guided image editing methods: Null-text Inversion~\cite{mokady2023null} combined with Prompt-to-Prompt~\cite{hertz2022prompt}, InstructPix2Pix~\cite{brooks2023instructpix2pix}, and SINE~\cite{SINE}.
InstructPix2Pix accepts editing instructions as its prompt inputs, such as ``Make her smile''.
In addition, we compare with a state-of-the-art GAN-based method DeltaEdit~\cite{lyu2023deltaedit}, which takes prompts inputs containing the word ``face'', \textit{e.g.}, ``Face with smile''.

\begin{figure*}[t]
  \centering
  \begin{subfigure}{0.4\linewidth}
    \includegraphics[width=1.0\linewidth]{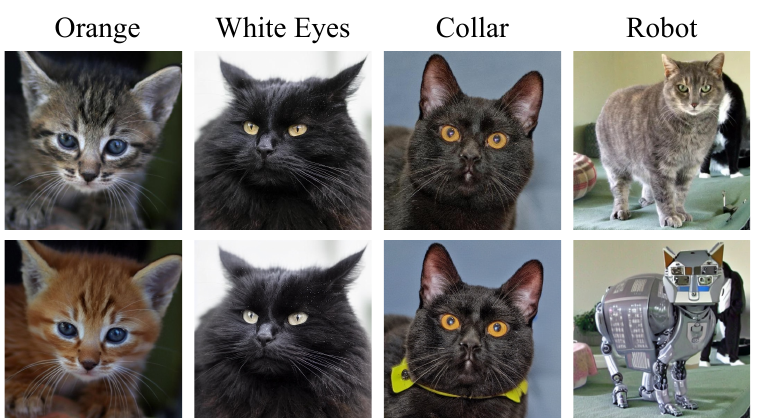}
  \end{subfigure}
  \hspace{0mm}
  \begin{subfigure}{0.4\linewidth}
    \includegraphics[width=1.0\linewidth]{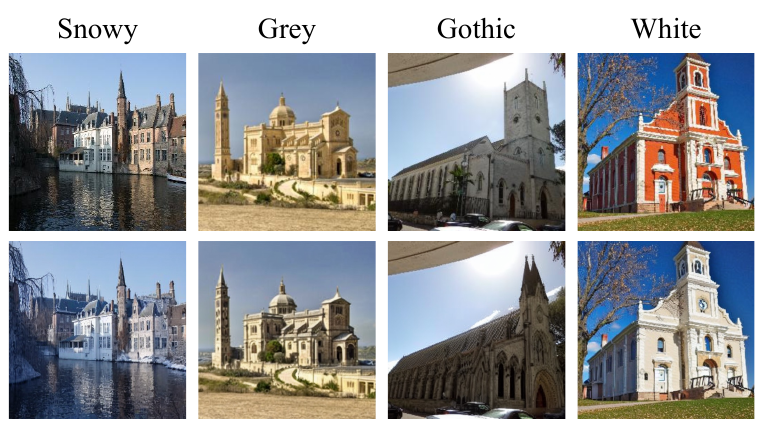}
  \end{subfigure}
  \caption{
    Editing results of \algoNameFull on cat (left) and church images (right).
  }
  \label{fig:cat-church-results}
  \vspace{-4mm}
\end{figure*}

\subsection{Qualitative Evaluation}\label{sec:qual-results}

\cref{fig:compare-baseline} shows the qualitative comparison between our method and four state-of-the-art methods on face editing. 
Our method faithfully preserves the original identities while accurately performing the desired edits. 
DeltaEdit~\cite{lyu2023deltaedit} follows the editing prompts well, but fails to preserve the identity of source images due to the limited capacity of GANs' latent space. 
For example, the hairstyle is changed in the top row, and the expression drifts in the fourth row. 
InsturctPix2Pix~\cite{brooks2023instructpix2pix} makes the edits but fails to locate the desired editing area, \textit{e.g.}, removing only part of the glasses in the third row, and colorizing the glasses which are undesired in the last two rows. 
Except for the eyeglass removal failure, 
Null-text Inversion~\cite{mokady2023null} struggles to balance identity preservation and prompt alignment.
For instance, the glasses are not removed in the third row, and identity drifts in the bottom row.
SINE~\cite{SINE} mostly produces unrealistic results with severe artifacts.
In \cref{fig:same-identity}, we experiment with different edits for the same human, showing versatile editing capability of \algoNameFull. 
Only the specified attribute of interest is manipulated, while irrelevant features are unaltered, which proves that \algoNameFull learns disentangled identity tokens. 
Overall, \algoNameFull achieves the most accurate editing results on human faces which requires precise control of localized details.

We further present manipulated results of non-face images such as cats and churches in \cref{fig:cat-church-results}, with editing targets ranging from global appearance to localized details.
\algoNameFull generates accurate and faithful editing results, indicating the generalizability of our approach to various domains.

\begin{figure}[t]
  \centering
  \includegraphics[width=1.0\linewidth]{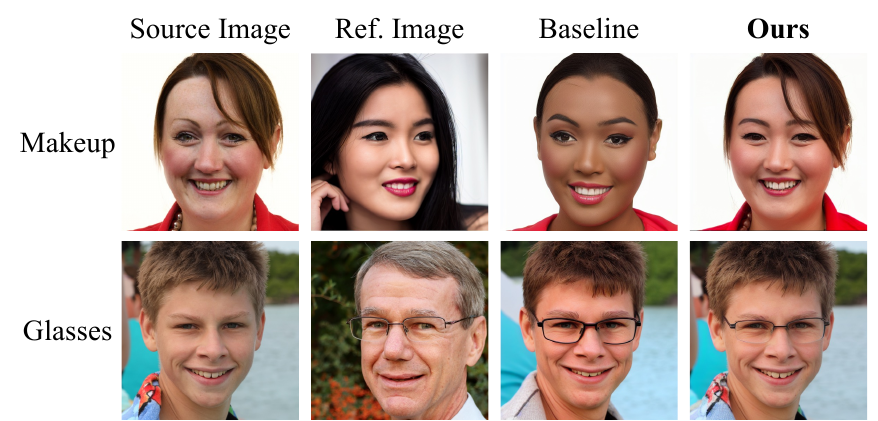}
  \caption{
    Compositional image editing results using \algoNameFull.
    We show facial attribute transfer applications from a reference (Ref.) image to a source image.
  }
  \label{fig:mix-results}
  \vspace{-5mm}
\end{figure}

The results of compositional image editing are shown in \cref{fig:mix-results}, where we transfer an attribute from the reference image to the source image. 
We compare \algoNameFull with a baseline method that only adopts plain identity fine-tuning.
Our method can extract the target attribute accurately and mix it with other facial features naturally, while keeping the person's identity intact. 
In contrast, the baseline method fails to balance identity preservation and attribute mixing, leading to changes in skin tones and distorted facial attributes.

See Appendix~\textcolor{red}{C} for more qualitative results of \algoNameFull.

\subsection{Quantitative Evaluation}\label{sec:quan-results}

\begin{table}[t]
  \centering
  \begin{tabular}{lccc}
    \toprule
    Method & FID ($\downarrow$) & LPIPS ($\downarrow$) & PSNR ($\uparrow$) \\
    \midrule
    Null-text Inversion & 67.61 & 0.18 & 30.29\\
    InstructPix2Pix & 56.98 & 0.15 & 30.48\\
    SINE & 107.56 & 0.38 & 28.56\\
    DeltaEdit & 86.41 & 0.30 & 29.01 \\
    \textbf{Ours} & \textbf{52.31} & \textbf{0.13} & \textbf{30.75}\\
    \bottomrule
  \end{tabular}
  \vspace{2mm}
  \caption{
    Quantitative results of image editing quality on FFHQ samples.
    \algoNameFull outperforms other methods across all metrics.
  }
  \label{tab:quant-results}
  \vspace{-6mm}
\end{table}


\begin{figure}[t]
  \centering
  \includegraphics[width=0.9\linewidth]{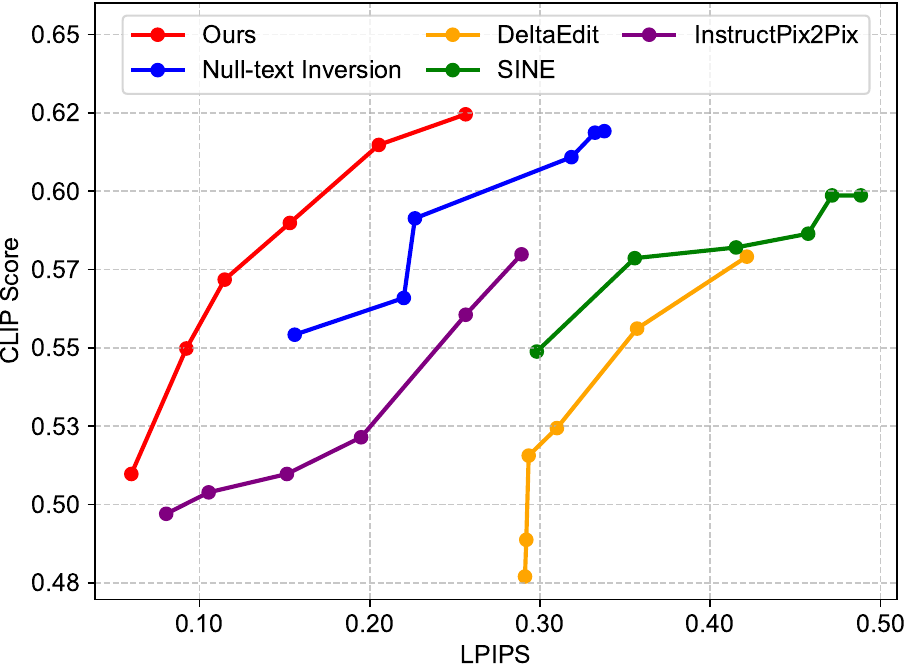}
  \caption{
    Trade-off between prompt alignment (CLIP score, higher means better alignment) and identity preservation (LPIPS, lower is better preservation) on FFHQ.
    For each method, we fix the prompts and vary the editing strength.
    \algoNameFull balances between the two properties the best.
  }
  \label{fig:clip-lpips-results}
  \vspace{-7mm}
\end{figure}

We report common metrics of all methods including FID~\cite{FID}, LPIPS~\cite{lpips}, and PSNR in \cref{tab:quant-results}. 
LPIPS and PSNR measure identity preservation, and FID reflects the image quality.
Following previous work~\cite{lyu2023deltaedit,kawar2023imagic}, the results are averaged over 150 samples, each with 10 editing prompts. 
\algoNameFull achieves the best results in all metrics, showing its superiority in editing quality.

In \cref{fig:clip-lpips-results}, we investigate the trade-off between prompt alignment and identity preservation measured by LPIPS, \textit{i.e.}, feature similarity between $\mathcal{I}$ and $\mathcal{I}^\ast$, and CLIP score~\cite{clipscore}, \textit{i.e.}, feature similarity between $\mathcal{P}^\ast$ and $\mathcal{I}^\ast$. 
LPIPS and CLIP score are competing metrics, since increasing the degree to which the edited images aligned with target prompts will reduce their similarity with the source images.
Still, \cref{fig:clip-lpips-results} shows that for a broad range of LPIPS values, our method has better prompt alignment for the same amount of image changes compared to other methods.

\begin{table}[h!]
  \centering
  \setlength{\tabcolsep}{4pt}
  \begin{tabular}{lcc}
    \toprule
    Method & ID. Preservation & Prompt Alignment  \\
    \midrule
    Null-text Inversion & 27.75\% & 26.00\% \\
    InstructPix2Pix & 33.13\% & 35.00\%\\
    SINE & 0.50\% & 10.75\% \\
    DeltaEdit & 30.00\% & 49.75\%  \\
    \textbf{Ours} & \textbf{71.38\%} & \textbf{72.38\%} \\
    \bottomrule
  \end{tabular}
  \vspace{2mm}
  \caption{
    User study on the identity (ID.) preservation and prompt alignment quality of editing results.
    We report preference rates of the two properties over \algoNameFull and other methods.
  }
  \label{tab:user-study}
  \vspace{-7mm}
\end{table}

\begin{figure}[h!]
  \centering
  \includegraphics[width=1.0\linewidth]{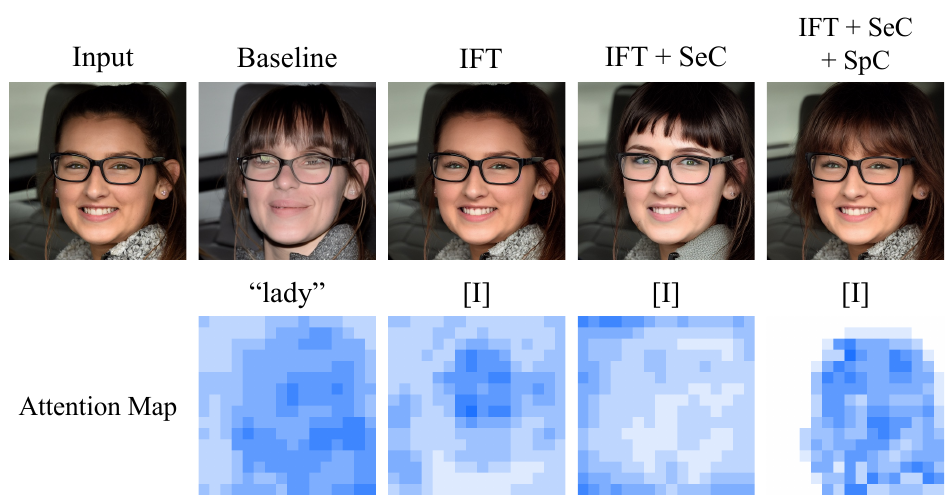}
  \caption{
    Analysis of each component of \algoNameFull.
    Top row: Edited images with different component(s) from \algoNameFull.
    Bottom row: Cross-attention maps of the word describing the person (``lady'' or [I]).
    IFT: Identity Fine-tuning, 
    SeC: Semantic Control, 
    SpC: Spatial Control.
    IFT + SeC + SpC is \algoNameFull.
  }
  \label{fig:ablation-attn}
  \vspace{-3mm}
\end{figure}

\begin{figure}[h!]
  \centering
  \includegraphics[width=1.0\linewidth]{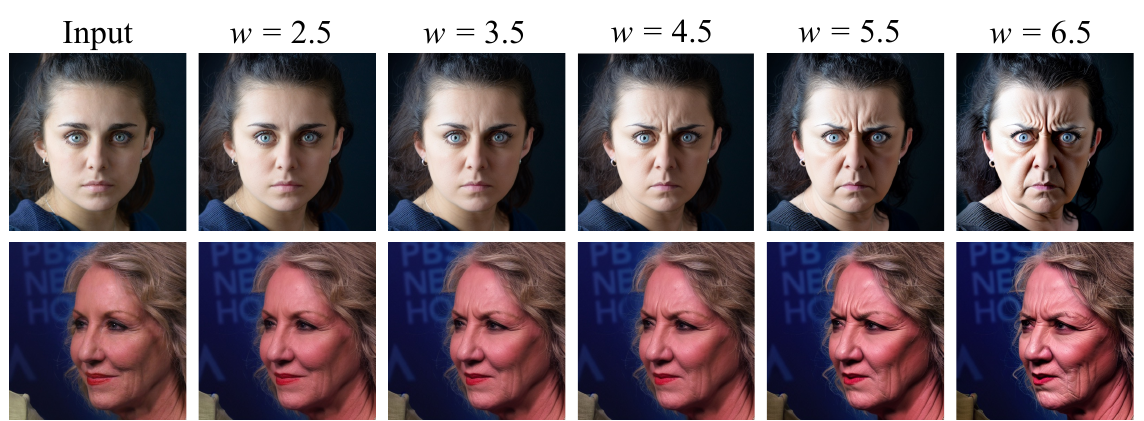}
  \caption{
    Analysis of classifier-free guidance scale $w$. 
    Images are edited with various $w$ and target prompt ``An angry lady''.
  }
  \label{fig:ablation-hyper}
  \vspace{-0mm}
\end{figure}

Finally, we conduct a user study on the identity preservation and prompt alignment of all methods. 
Over 40 questions involving 40 participants, we assessed the performance of different methods.
Participants evaluated randomly chosen pairs of original and edited images from \algoNameFull and four state-of-the-art methods, choosing those that best preserved identity and aligned with prompts (users can select more than one entry for each question). 
The results, detailed in \cref{tab:user-study}, reveal that \algoNameFull outperforms others in both properties.

\subsection{Ablation Analysis}\label{sec:ablations}

\heading{Effect of Each Component.} 
In \cref{fig:ablation-attn}, we analyze the effect of all three components within \algoNameFull: Identity Fine-tuning (IFT), Semantic Control (SeC), and Spatial Control (SpC).
Without them, our method falls back to the Null-text Inversion~\cite{mokady2023null} baseline, where the person's identity information is completely lost in the edited image.
Although this issue can be resolved by naively applying Identity Fine-tuning to learn an identity token [I], which fundamentally aligns with DreamBooth~\cite{ruiz2023dreambooth} that preserves the person's appearance, the target attribute included in the prompt (``bangs") is missing in the edited image.
The attention map of [I] reveals that this is caused by excessive information being encoded in [I], which limits the editability of the person's hair.
Adding Semantic Control helps alleviate this problem by disentangling [I] from information contained in the original prompt.
However, the lack of spatial constraints on the attention map causes [I]'s attention to drift towards irrelevant regions such as the background, while focusing less on the person's face.
The generated image under Identity Fine-tuning and Semantic Control aligns with the target prompt well, yet it distorts subtle details like the person's skin tone.
Finally, by adding Spatial Control that exclusively directs [I]'s attention to the region of interest, the edited image from our full method not only accurately follows the target prompt to add bangs, but also keeps the person's identity intact.

\heading{Impact of Guidance Scale.}
In \cref{fig:ablation-hyper}, we analyze the impact of guidance scale $w$ on the edited image.
Prompt alignment and identity preservation are two competing goals in image editing.
A higher guidance scale $w$ will induce more pronounced alterations in the direction guided by the editing prompt, albeit at the expense of losing identity information.
Empirically, $w$ within the range of $[3.5, 5.5]$ leads to balanced results.

\section{Conclusion}\label{sec:conclusion}

We propose \algoNameFull, a novel text-guided image editing method.
Our method executes precise edits while preserving the original identity intact.
This is achieved by fine-tuning a pre-trained text-to-image diffusion model with semantic and spatial controls.
To apply accurate and faithful edits, we insert a learnable token and constrain its textual feature space and spatial focus during fine-tuning.
Extensive evaluations show \algoNameFull's superiority over state-of-the-art methods qualitatively and quantitatively.
Moreover, \algoNameFull's adaptability to compositional image editing illustrates its flexibility.

\heading{Limitations.}
A text prompt describing the source image is required for editing, which is a common constraint in methods building on pre-trained text-to-image diffusion models~\cite{hertz2022prompt,SINE}.
A future direction is thus to enable editing without source prompts, which may be achieved by inverting prompts from images~\cite{PromptTuningInversion} or captioning models~\cite{li2022blip}.

\section*{Impact Statement}
\algoNameFull makes powerful image editing accessible for positive use cases such as creative production and education, but the same capability can also be misused to create deceptive or non-consensual content. Responsible deployment therefore requires provenance tools (e.g., watermarking), clear usage policies, and continued auditing for bias and misuse.

\bibliographystyle{icml2025}
\bibliography{reference}

@String(PAMI  = {IEEE Trans. Pattern Anal. Mach. Intell.})

@String(CVPR  = {IEEE Conf. Comput. Vis. Pattern Recog.})

@String(ICCV  = {Int. Conf. Comput. Vis.})

@String(ECCV  = {Eur. Conf. Comput. Vis.})

@String(NeurIPS = {Adv. Neural Inform. Process. Syst.})

@String(ICML  = {Int. Conf. Mach. Learn.})

@String(ICLR  = {Int. Conf. Learn. Represent.})

@String(TOG   = {ACM Trans. Graph.})

@String(PAMI  = {IEEE TPAMI})

@String(CVPR  = {CVPR})

@String(ICCV  = {ICCV})

@String(ECCV  = {ECCV})

@String(NeurIPS = {NeurIPS})

@String(ICML  = {ICML})

@String(ICLR  = {ICLR})

@String(TOG   = {ACM TOG})

@article{adamw,
  title={Decoupled weight decay regularization},
  author={Loshchilov, Ilya and Hutter, Frank},
  journal={arXiv preprint arXiv:1711.05101},
  year={2017}
}

@inproceedings{paint-by-example,
  title={Paint by example: Exemplar-based image editing with diffusion models},
  author={Yang, Binxin and Gu, Shuyang and Zhang, Bo and Zhang, Ting and Chen, Xuejin and Sun, Xiaoyan and Chen, Dong and Wen, Fang},
  booktitle={CVPR},
  year={2023}
}

@inproceedings{afhq,
  title={{StarGAN} v2: Diverse image synthesis for multiple domains},
  author={Choi, Yunjey and Uh, Youngjung and Yoo, Jaejun and Ha, Jung-Woo},
  booktitle={CVPR},
  year={2020}
}

@article{clipscore,
  title={{CLIPScore}: A reference-free evaluation metric for image captioning},
  author={Hessel, Jack and Holtzman, Ari and Forbes, Maxwell and Bras, Ronan Le and Choi, Yejin},
  journal={arXiv preprint arXiv:2104.08718},
  year={2021}
}

@inproceedings{lpips,
  title={The unreasonable effectiveness of deep features as a perceptual metric},
  author={Zhang, Richard and Isola, Phillip and Efros, Alexei A and Shechtman, Eli and Wang, Oliver},
  booktitle={CVPR},
  year={2018}
}

@inproceedings{FID,
  title={{GANs} trained by a two time-scale update rule converge to a local nash equilibrium},
  author={Heusel, Martin and Ramsauer, Hubert and Unterthiner, Thomas and Nessler, Bernhard and Hochreiter, Sepp},
  booktitle={NeurIPS},
  year={2017}
}

@article{lsun,
  title={{LSUN}: Construction of a large-scale image dataset using deep learning with humans in the loop},
  author={Yu, Fisher and Seff, Ari and Zhang, Yinda and Song, Shuran and Funkhouser, Thomas and Xiao, Jianxiong},
  journal={arXiv preprint arXiv:1506.03365},
  year={2015}
}

@inproceedings{celeba,
  title = {Deep Learning Face Attributes in the Wild},
  author = {Liu, Ziwei and Luo, Ping and Wang, Xiaogang and Tang, Xiaoou},
  booktitle = {ICCV},
  year = {2015} 
}

@inproceedings{radford2021learning,
  title={Learning transferable visual models from natural language supervision},
  author={Radford, Alec and Kim, Jong Wook and Hallacy, Chris and Ramesh, Aditya and Goh, Gabriel and Agarwal, Sandhini and Sastry, Girish and Askell, Amanda and Mishkin, Pamela and Clark, Jack and others},
  booktitle={ICML},
  year={2021}
}

@inproceedings{li2022blip,
  title={{BLIP}: Bootstrapping language-image pre-training for unified vision-language understanding and generation},
  author={Li, Junnan and Li, Dongxu and Xiong, Caiming and Hoi, Steven},
  booktitle={ICML},
  year={2022}
}

@inproceedings{sohl2015deep,
  title={Deep unsupervised learning using nonequilibrium thermodynamics},
  author={Sohl-Dickstein, Jascha and Weiss, Eric and Maheswaranathan, Niru and Ganguli, Surya},
  booktitle={ICML},
  year={2015}
}

@inproceedings{song2020score,
  title={Score-Based Generative Modeling through Stochastic Differential Equations},
  author={Song, Yang and Sohl-Dickstein, Jascha and Kingma, Diederik P and Kumar, Abhishek and Ermon, Stefano and Poole, Ben},
  booktitle={ICLR},
  year={2020}
}

@inproceedings{ho2020denoising,
  title={Denoising diffusion probabilistic models},
  author={Ho, Jonathan and Jain, Ajay and Abbeel, Pieter},
  booktitle={NeurIPS},
  year={2020}
}

@inproceedings{nichol2021improved,
  title={Improved denoising diffusion probabilistic models},
  author={Nichol, Alexander Quinn and Dhariwal, Prafulla},
  booktitle={ICML},
  year={2021}
}

@inproceedings{dhariwal2021diffusion,
  title={Diffusion models beat {GANs} on image synthesis},
  author={Dhariwal, Prafulla and Nichol, Alexander},
  booktitle={NeurIPS},
  year={2021}
}

@inproceedings{rombach2022high,
  title={High-resolution image synthesis with latent diffusion models},
  author={Rombach, Robin and Blattmann, Andreas and Lorenz, Dominik and Esser, Patrick and Ommer, Bj{\"o}rn},
  booktitle={CVPR},
  year={2022}
}

@inproceedings{saharia2022photorealistic,
  title={Photorealistic text-to-image diffusion models with deep language understanding},
  author={Saharia, Chitwan and Chan, William and Saxena, Saurabh and Li, Lala and Whang, Jay and Denton, Emily L and Ghasemipour, Kamyar and Gontijo Lopes, Raphael and Karagol Ayan, Burcu and Salimans, Tim and others},
  booktitle={NeurIPS},
  year={2022}
}

@article{unCLIP,
  title={Hierarchical text-conditional image generation with {CLIP} latents},
  author={Ramesh, Aditya and Dhariwal, Prafulla and Nichol, Alex and Chu, Casey and Chen, Mark},
  journal={arXiv preprint arXiv:2204.06125},
  year={2022}
}

@inproceedings{nichol2021glide,
  title={{GLIDE}: Towards Photorealistic Image Generation and Editing with Text-Guided Diffusion Models},
  author={Nichol, Alexander Quinn and Dhariwal, Prafulla and Ramesh, Aditya and Shyam, Pranav and Mishkin, Pamela and Mcgrew, Bob and Sutskever, Ilya and Chen, Mark},
  booktitle={ICML},
  year={2022}
}

@inproceedings{meng2021sdedit,
  title={{SDEdit}: Guided Image Synthesis and Editing with Stochastic Differential Equations},
  author={Meng, Chenlin and He, Yutong and Song, Yang and Song, Jiaming and Wu, Jiajun and Zhu, Jun-Yan and Ermon, Stefano},
  booktitle={ICLR},
  year={2021}
}

@inproceedings{BlendedDMEdit,
  title={Blended diffusion for text-driven editing of natural images},
  author={Avrahami, Omri and Lischinski, Dani and Fried, Ohad},
  booktitle={CVPR},
  year={2022}
}

@article{BlendedLDMEdit,
  title={Blended latent diffusion},
  author={Avrahami, Omri and Fried, Ohad and Lischinski, Dani},
  journal={TOG},
  year={2023}
}

@inproceedings{kim2022diffusionclip,
  title={{DiffusionCLIP}: Text-guided diffusion models for robust image manipulation},
  author={Kim, Gwanghyun and Kwon, Taesung and Ye, Jong Chul},
  booktitle={CVPR},
  year={2022}
}

@inproceedings{bansal2023universal,
  title={Universal guidance for diffusion models},
  author={Bansal, Arpit and Chu, Hong-Min and Schwarzschild, Avi and Sengupta, Soumyadip and Goldblum, Micah and Geiping, Jonas and Goldstein, Tom},
  booktitle={CVPR},
  year={2023}
}

@inproceedings{hertz2022prompt,
  title={Prompt-to-Prompt Image Editing with Cross-Attention Control},
  author={Hertz, Amir and Mokady, Ron and Tenenbaum, Jay and Aberman, Kfir and Pritch, Yael and Cohen-or, Daniel},
  booktitle={ICLR},
  year={2022}
}

@inproceedings{PnP-Diffusion,
  title={Plug-and-play diffusion features for text-driven image-to-image translation},
  author={Tumanyan, Narek and Geyer, Michal and Bagon, Shai and Dekel, Tali},
  booktitle={CVPR},
  year={2023}
}

@inproceedings{parmar2023zero,
  title={Zero-shot image-to-image translation},
  author={Parmar, Gaurav and Kumar Singh, Krishna and Zhang, Richard and Li, Yijun and Lu, Jingwan and Zhu, Jun-Yan},
  booktitle={SIGGRAPH},
  year={2023}
}

@inproceedings{SINE,
  title={{SINE}: Single image editing with text-to-image diffusion models},
  author={Zhang, Zhixing and Han, Ligong and Ghosh, Arnab and Metaxas, Dimitris N and Ren, Jian},
  booktitle={CVPR},
  year={2023}
}

@inproceedings{brooks2023instructpix2pix,
  title={{InstructPix2Pix}: Learning to follow image editing instructions},
  author={Brooks, Tim and Holynski, Aleksander and Efros, Alexei A},
  booktitle={CVPR},
  year={2023}
}

@inproceedings{mokady2023null,
  title={Null-text inversion for editing real images using guided diffusion models},
  author={Mokady, Ron and Hertz, Amir and Aberman, Kfir and Pritch, Yael and Cohen-Or, Daniel},
  booktitle={CVPR},
  year={2023}
}

@inproceedings{kawar2023imagic,
  title={Imagic: Text-based real image editing with diffusion models},
  author={Kawar, Bahjat and Zada, Shiran and Lang, Oran and Tov, Omer and Chang, Huiwen and Dekel, Tali and Mosseri, Inbar and Irani, Michal},
  booktitle={CVPR},
  year={2023}
}

@inproceedings{PromptTuningInversion,
  title={Prompt Tuning Inversion for Text-Driven Image Editing Using Diffusion Models},
  author={Dong, Wenkai and Xue, Song and Duan, Xiaoyue and Han, Shumin},
  booktitle={ICCV},
  year={2023}
}

@inproceedings{pan2023effective,
  title={Effective Real Image Editing with Accelerated Iterative Diffusion Inversion},
  author={Pan, Zhihong and Gherardi, Riccardo and Xie, Xiufeng and Huang, Stephen},
  booktitle={ICCV},
  year={2023}
}

@inproceedings{DM-GAN-Edit,
  title={Not All Steps are Created Equal: Selective Diffusion Distillation for Image Manipulation},
  author={Wang, Luozhou and Yang, Shuai and Liu, Shu and Chen, Ying-cong},
  booktitle={ICCV},
  year={2023}
}

@inproceedings{gal2022image,
  title={An Image is Worth One Word: Personalizing Text-to-Image Generation using Textual Inversion},
  author={Gal, Rinon and Alaluf, Yuval and Atzmon, Yuval and Patashnik, Or and Bermano, Amit Haim and Chechik, Gal and Cohen-or, Daniel},
  booktitle={ICLR},
  year={2022}
}

@inproceedings{ruiz2023dreambooth,
  title={{DreamBooth}: Fine tuning text-to-image diffusion models for subject-driven generation},
  author={Ruiz, Nataniel and Li, Yuanzhen and Jampani, Varun and Pritch, Yael and Rubinstein, Michael and Aberman, Kfir},
  booktitle={CVPR},
  year={2023}
}

@inproceedings{kumari2023multi,
  title={Multi-concept customization of text-to-image diffusion},
  author={Kumari, Nupur and Zhang, Bingliang and Zhang, Richard and Shechtman, Eli and Zhu, Jun-Yan},
  booktitle={CVPR},
  year={2023}
}

@article{shi2023instantbooth,
  title={{InstantBooth}: Personalized text-to-image generation without test-time finetuning},
  author={Shi, Jing and Xiong, Wei and Lin, Zhe and Jung, Hyun Joon},
  journal={arXiv preprint arXiv:2304.03411},
  year={2023}
}

@inproceedings{goodfellow2014generative,
  title={Generative adversarial nets},
  author={Goodfellow, Ian and Pouget-Abadie, Jean and Mirza, Mehdi and Xu, Bing and Warde-Farley, David and Ozair, Sherjil and Courville, Aaron and Bengio, Yoshua},
  booktitle={NeurIPS},
  year={2014}
}

@inproceedings{GANEdit1,
  title={Semantic image synthesis via adversarial learning},
  author={Dong, Hao and Yu, Simiao and Wu, Chao and Guo, Yike},
  booktitle={ICCV},
  year={2017}
}

@inproceedings{GANEdit2,
  title={{ManiGAN}: Text-guided image manipulation},
  author={Li, Bowen and Qi, Xiaojuan and Lukasiewicz, Thomas and Torr, Philip HS},
  booktitle={CVPR},
  year={2020}
}

@inproceedings{GANEdit3,
  title={Text-adaptive generative adversarial networks: manipulating images with natural language},
  author={Nam, Seonghyeon and Kim, Yunji and Kim, Seon Joo},
  booktitle={NeurIPS},
  year={2018}
}

@inproceedings{karras2019style,
  title={A style-based generator architecture for generative adversarial networks},
  author={Karras, Tero and Laine, Samuli and Aila, Timo},
  booktitle={CVPR},
  year={2019}
}

@inproceedings{karras2020analyzing,
  title={Analyzing and improving the image quality of {StyleGAN}},
  author={Karras, Tero and Laine, Samuli and Aittala, Miika and Hellsten, Janne and Lehtinen, Jaakko and Aila, Timo},
  booktitle={CVPR},
  year={2020}
}

@inproceedings{karras2021alias,
  title={Alias-free generative adversarial networks},
  author={Karras, Tero and Aittala, Miika and Laine, Samuli and H{\"a}rk{\"o}nen, Erik and Hellsten, Janne and Lehtinen, Jaakko and Aila, Timo},
  booktitle={NeurIPS},
  year={2021}
}

@inproceedings{wu2021stylespace,
  title={{Stylespace Analysis}: Disentangled controls for {StyleGAN} image generation},
  author={Wu, Zongze and Lischinski, Dani and Shechtman, Eli},
  booktitle={CVPR},
  year={2021}
}

@article{shen2020interfacegan,
  title={{InterFaceGAN}: Interpreting the disentangled face representation learned by {GANs}},
  author={Shen, Yujun and Yang, Ceyuan and Tang, Xiaoou and Zhou, Bolei},
  journal={PAMI},
  year={2020}
}

@inproceedings{harkonen2020ganspace,
  title={{GANSpace}: Discovering interpretable {GAN} controls},
  author={H{\"a}rk{\"o}nen, Erik and Hertzmann, Aaron and Lehtinen, Jaakko and Paris, Sylvain},
  booktitle={NeurIPS},
  year={2020}
}

@inproceedings{chen2022exploring,
  title={Exploring Gradient-Based Multi-directional Controls in {GANs}},
  author={Chen, Zikun and Jiang, Ruowei and Duke, Brendan and Zhao, Han and Aarabi, Parham},
  booktitle={ECCV},
  year={2022}
}

@inproceedings{richardson2021encoding,
  title={Encoding in style: a {StyleGAN} encoder for image-to-image translation},
  author={Richardson, Elad and Alaluf, Yuval and Patashnik, Or and Nitzan, Yotam and Azar, Yaniv and Shapiro, Stav and Cohen-Or, Daniel},
  booktitle={CVPR},
  year={2021}
}

@article{tov2021designing,
  title={Designing an encoder for {StyleGAN} image manipulation},
  author={Tov, Omer and Alaluf, Yuval and Nitzan, Yotam and Patashnik, Or and Cohen-Or, Daniel},
  journal={TOG},
  year={2021}
}

@inproceedings{wang2022high,
  title={High-fidelity {GAN} inversion for image attribute editing},
  author={Wang, Tengfei and Zhang, Yong and Fan, Yanbo and Wang, Jue and Chen, Qifeng},
  booktitle={CVPR},
  year={2022}
}

@inproceedings{lyu2023deltaedit,
  title={{DeltaEdit}: Exploring text-free training for text-driven image manipulation},
  author={Lyu, Yueming and Lin, Tianwei and Li, Fu and He, Dongliang and Dong, Jing and Tan, Tieniu},
  booktitle={CVPR},
  year={2023}
}


\end{document}